# Seed Phenotyping on Neural Networks using Domain Randomization and Transfer Learning

Venkat Margapuri and Mitchell Neilsen

Department of Computing and Information Sciences

Kansas State University, Manhattan, KS

**ABSTRACT**

Seed phenotyping is the idea of analyzing the morphometric characteristics of a seed to predict the behavior of the seed in terms of development, tolerance and yield in various environmental conditions. The focus of the work is the application and feasibility analysis of the state-of-the-art object detection and localization neural networks, Mask R-CNN and YOLO (You Only Look Once), for seed phenotyping using Tensorflow. One of the major bottlenecks of such an endeavor is the need for large amounts of training data. While the capture of a multitude of seed images is taunting, the images are also required to be annotated to indicate the boundaries of the seeds on the image and converted to data formats that the neural networks are able to consume. Although tools to manually perform the task of annotation are available for free, the amount of time required is enormous. In order to tackle such a scenario, the idea of domain randomization i.e. the technique of applying models trained on images containing simulated objects to real-world objects, is considered. In addition, transfer learning i.e. the idea of applying the knowledge obtained while solving a problem to a different problem, is used. The networks are trained on pre-trained weights from the popular ImageNet and COCO data sets. A technique that closely follows the guidelines laid out by the International Seed Morphology Association is proposed to estimate the morphometry of the seeds using the neural networks.

## INTRODUCTION

The developments in the areas of computer vision and deep learning in recent times have helped computer scientists conduct seed phenotyping and make portable solutions available to plant scientists aiding them in their research endeavors. One such effort is this work that utilizes the state-of-the-art object detection neural networks, Mask R-CNN and YOLO (You Only Look Once) to propose a neural network solution for seed phenotyping. The use of neural networks requires an extensive amount of training data that can be hard to acquire for various reasons. To name a few, the availability of seeds, capture of several images containing seeds, annotation of each of the captured images with bounding boxes and finally, conversion of the annotations into data formats that are consumed by the neural networks. In order to alleviate the painstaking process of training data creation, Domain Randomization i.e. the technique of training models on simulated images that transfer to real images, is applied. The application of this technique means that a high number of seed samples is not necessary and a variety of images containing simulated seeds in different sizes and orientations that match real seeds can be generated. Another aspect of neural networks that makes them good candidates for seed phenotyping is their ability to learn from models previously trained on other data sets, typically known as Transfer Learning. While several popular pre-trained models (weights) are available for transfer learning, the weights from COCO and ImageNet data sets for Mask R-CNN and YOLOv5, and Tiny-YOLOv4 weights for the Tiny-YOLOv4 model are used for experiments in this work. In addition to a seed detection technique, the work also proposes the estimation of seed morphometry with the help of a ground-truth, the US Penny. The obtained results are compared against prevailing seed morphometry estimation applications, Seed Counter (Komyshev, 2017), Leaf-IT (Schrader et al., 2017) and the Mesh (Margapuri et al., 2020) algorithm. The results show that the estimated morphometry is comparable to that of the prevailing applications.

In the remainder of the article, section 2 mentions related work, section 3 discusses materials and methods required for the implementation of the project, section 4 walks through the estimation of seed morphometry and section 5 concludes the work.

## 1. Related Work

In the realm of domain randomization and synthetic data generation, 'Training Instance Segmentation Neural Networks with Synthetic Data sets for Crop Seed Phenotyping' by Yoda et al. discusses the application of synthetic data sets to Mask R-CNN for seed phenotyping and is most relevant to the current work. Similarly, 'Data Augmentation for Leaf Segmentation and Counting Tasks in Rosette Plants' by Kuzinchov et. Al proposes a technique named 'Collage' that applies the idea of synthetic data sets to images of leaves. From a morphometric estimation standpoint, the mobile applications SmartGrain (Tanabata et al., 2012), LeafIT (Schrader et al., 2017), Seed Counter (Komyshev, 2017) and the algorithms Mesh (Margapuri et al., 2020) and No-Mesh (Margapuri et al., 2020) served as the basis for implementation and comparison.

## 2. MATERIALS AND METHODS

### 2.1 Image Acquisition

Initially, the images for each of the seeds in the considered sample are required to be acquired. For the purposes of this experiment, 40 seeds are considered for each seed type of which 25 are used for training/ validation and 15 for testing. For the coins, 10 pennies are considered of which seven are used for training and three for testing. Fewer pennies are selected since pennies are uniform as opposed to seeds having random shapes. The images are captured using a Moto G6 mobile phone placed on a 3-D stand that holds the phone orthogonal to the ground so as to eliminate any skew that might result from holding the phone free-hand. For the background, a lightbox emitting white light is chosen. The resolution of the captured images is 3456 x 4608 px. Figure 1 shows the capture of the 25 seed sample of soy selected for training. Likewise, images of the others are captured.

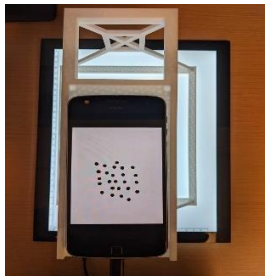

**Figure 1:** Image Acquisition of Soy Seeds

## 2.2 Synthetic Datasets

As part of the work, synthetic data sets for six different entities, five types of seeds and one type of coins are created. The seeds used are canola, rough rice, sorghum, soy and wheat. The coin used is a US Penny. In either of the neural networks i.e. Mask R-CNN or YOLO, the training images are required to be annotated. The creation of data sets can be done manually using tools such as Oxford's VGG Image Annotator that provide a fluid UI to annotate the images. However, such an annotation procedure is cumbersome and laborious. In order to efficiently perform the creation and annotation of training images, a technique named Domain Randomization i.e. the idea of training models on images containing simulated objects that translate closely to real-world objects, is applied. The application of domain randomization calls for a subset of the entity population to represent the entire population of the entity. As mentioned earlier, 40 samples of each seed type and 10 pennies are selected as the entities to represent the entire population. The steps in the application of domain randomization upon the selection of representative samples are as follows:

1. Capture images of each of the samples as shown in Figure 1 and extract out the seeds (or coins) on the image. Typically, the seeds (or coins) are the foreground on the image and are extracted by separating the alpha channel from the image. Such an extraction can be performed using tools such as Gimp or OpenCV GrabCut. For this work, Gimp is used.

2. Select a canvas onto which the extracted samples can be laid. The canvas functions as the background of the image and it is recommended that a canvas that resembles the real-world test site be picked. If there are different test sites, it is best to have at least one image of each of the test sites.

3. Now that foregrounds and backgrounds are available, select all samples belonging to a particular foreground and lay them on each of the backgrounds with varying sizes and orientations. Other kinds of augmentation such as brightness are also be performed as needed.

4. As the images are being created in step 3, concurrently create a black canvas and color each of the regions where the seed (or coin) is placed on the image. Use a unique color for each of the objects on the image so as to identify them later.

Following the described approach, multiple synthetic data sets containing 275 images of each of the seeds and pennies with each image having between 450 and 600 objects are created for training. Note that fewer objects i.e. between 50 and 100 for images containing pennies are used since they are uniform in size. The intuition behind the data sets is to have them resemble the real-world images as closely as possible. In a real-world image, the instances on the images touch and segmenting such instances is known to be a hard problem to solve. In order to evaluate the ability of the neural networks to perform such a segmentation, the seeds on the training images are allowed to touch each other and pile up on occasion. Earlier work by Yoda et al. does not allow for more than a 25% overlap of the instances as the images are generated. As part of the current work, no bounds on overlap are set as part of data set creation since the large number of images for each seed type means that there a sufficient number of training samples with different degrees of overlap. Figure 2 shows a sample of each of the seeds and their corresponding masks with varying degrees of overlap.

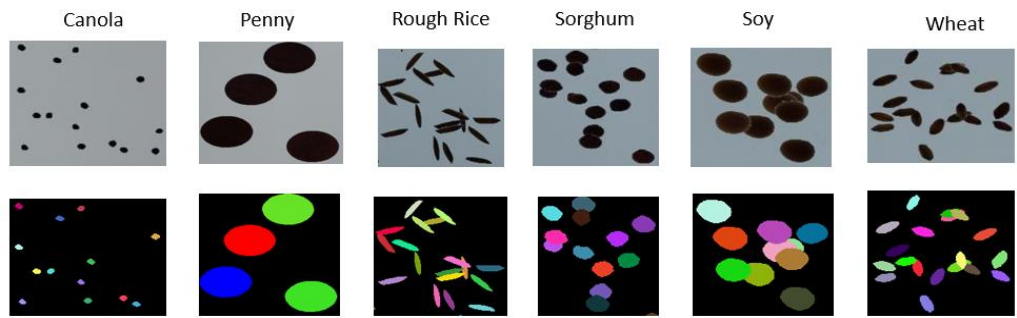

**Figure 2:** Sample of Seeds and Corresponding Masks for Synthetic Data Sets

## 2.2 Image Annotations

Once the images are generated using the aforementioned approach, the next step is to annotate each of the instances in the images. Annotations, simply, are files that contain information about the image ID, boundary, class, area and any other characteristic of each of the objects in the image. While information consumed by either of the neural networks is similar in nature, the format in which the information is consumed differs. Mask R-CNN consumes annotations in the COCO file format whereas YOLO consumes information in the TXT file format. Using the earlier work by Yehao, the generation of annotations is automated. In addition, an online tool named Roboflow comes handy to generate annotations. The tool is a pay-to-play model after the first 1000 images and so, helps users that are either willing to pay or have fewer than 1000 images to annotate.

## 2.3 Mask RCNN

The Mask RCNN implementation used is the popular implementation by Matterport that runs on Tensorflow. Mask RCNN comes pretrained on 80 classes that belong to the COCO data set. For a majority of the applications, the default parameters shipped out by the Matterport implementation are good enough for object detection. However, for the object detection of the seeds and pennies, one key change that is made is the 'anchor scales' and 'anchor ratios' of the Region Proposal Network. Anchor scales and anchor ratios play a crucial role in instance detection. For instance, say the network is trained on canola with images of size 768 x 768 px. If the anchor scales were selected to be [32, 64] and anchor ratios, which are essentially aspect ratios, were [1, 2], four anchor boxes would be generated at each position with dimensions 32:32 px, ~22:44 px, 64:64 px, ~90:45 px. However, from experimentation, it is observed that even the largest of canola seeds don't occupy nearly as close to the smallest of the anchors 32:32 px (or 22:44 px). As a result, the network fails to train well yielding a higher overall loss and sloppy detection. In order to ensure this does not happen, different anchor scales and ratios are experimented with and the values of [2, 4, 16, 32, 64] and [1.5, 2, 3] are settled on for anchor scales and ratios respectively. For this work, two different models are created using the pre-trained weights of COCO and ImageNet with the backbone of ResNet101 architecture. While ImageNet is a data set that comprises of 1000 classes in comparison to COCO's 80, it is worth noting that neither ImageNet nor COCO was trained on any kind of seeds originally. While the performance of both models is similar, there is one key fallacy that they possess pertinent to seed shape. Amongst the five seeds in question, the seeds of soy and sorghum are circular and similar in size on occasion. Likewise, wheat and rough rice, except that they are oblong in shape. As a consequence, both models tend to commonly mistake rough rice for wheat and vice-versa, and soy for sorghum and vice-versa. The seeds of canola, although circular like soy and sorghum are significantly smaller in comparison leading to a lower error rate while detecting them. Figure 3 shows the detections of seeds on each of the images in question. While the classification of seeds may not be accurate, a point worth noting is that the instance segmentation performed by the networks is beneficial by itself because the majority of applications, desktop or mobile, are catered to work for a certain type of seed. As a result, classification of seeds isn't as critical as instance segmentation.

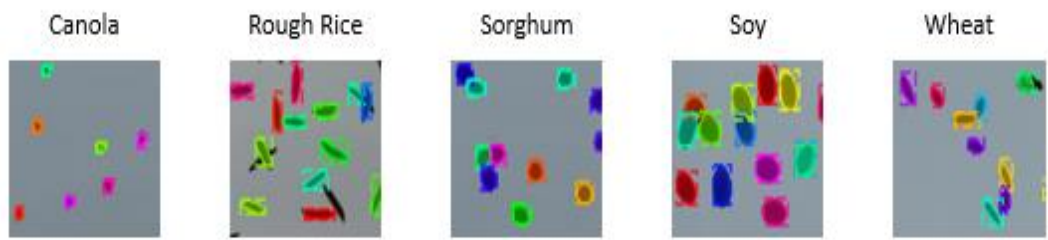

**Figure 3:** Mask RCNN Instance Detections on Different Seeds

### 2.4.1 YOLO

You Only Look Once (YOLO) is one of the first object detection models that introduced the idea of combining bounding box prediction and object classification into a single network. The architecture got introduced as part of a framework called Darknet. YOLO has come out with timely iterations with the latest iteration being five. Each of the iterations, except the fifth iteration has architectural changes done to improve upon the performance of the model. YOLOv5 is the architecture that is considered for experimentation as part of the work. YOLOv5 consists of four different architectures, YOLOv5s, YOLOv5m, YOLOv5l and YOLOv5x. Each of the architectures differ in the number of neural network layers with 5s, 5m, 5l and 5x containing 283, 391, 499 and 607 layers respectively. All four of the official models made available are trained from scratch on the COCO dataset. As part of the work, all four architectures are experimented.

The observation is that the performance of the network improves as the number of layers increases. Figure 3 shows the detections made by YOLOv5 on the different entities that the model is trained.

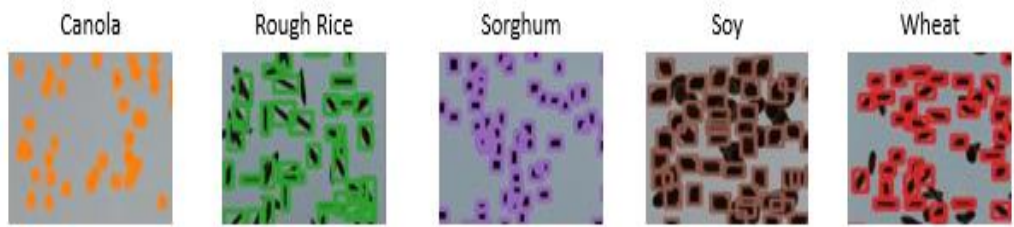

**Figure 4:** YOLOv5x Instance Detections on Clustered Seeds

## 2.4.2 Tiny-YOLO

The models created using YOLOv5 are too complex and large to deploy natively as mobile apps. Tiny-YOLO models are a variation of the YOLO models but are much smaller in comparison to the YOLO models. However, the models are not accurate as their bigger counterparts. For the purposes of this work, YOLOv4-Tiny is considered since it is closest to the YOLOv5 models at the time of experimentation. The average precision (AP) at 50% mask-IOU of YOLOv4-Tiny is 40.2% in comparison to YOLOv4's 64.9% on the COCO data set. In exchange for accuracy, the speed of the models has a multifold increase. The Tiny-YOLOv4 achieves about 371 FPS during inference on a single GPU (GTX 1080 Ti) in comparison to YOLOv4's 62 FPS on a single GPU (TeslaV100) on the Darknet framework and is about six times faster. While there is a significant decrease in the accuracy of the Tiny-YOLO models in comparison to YOLO models on the complex COCO data set, it is worth noting that Tiny-YOLO models function nearly as efficiently as the larger YOLO models when trained on custom data sets that are not as complex, like the ones in this work. A key difference between the larger models and Tiny-YOLOv4 is that the input image resolution of Tiny-YOLOv4 is 416 x416 px as opposed to the larger models' 768 x768 px.

## 3. RESULTS AND DISCUSSION

The project starts off with work on Mask RCNN where the network, once with pre-trained weights of COCO and another time with pre-trained weights of ImageNet, is trained on 31 images of soy where each image contains multiple (about 20 - 30) seeds. The images are manually annotated using Oxford's VGG Annotator tool. Although the network trains well and achieves a decent loss of ~2, the performance of the model on the test data set is sub-par. It is after the experience that the use of domain randomization is employed. The trained models are tested on two test datasets, one that contains real-world images of seeds and another of synthesized seed images. One of the criticisms of the work by Yoda et. al is that the metric scores represented for the synthetic data set could be biased since the synthetic test data set is created using the same seed samples used for training. In order to test out the alternative, the synthetic test data set is created using different seed samples. Overall, 150 images of size 768 x 768 px (416 x 416 px for YOLO-Tinyv4) are created where each type of seed contains 15 synthetic images and 15 real-world images containing 50 - 100 seeds. For the real-world test data set on each seed, about 5 of the images do not contain entities that touch each other but the remainder of the images do. The reason for such a setup is to observe the performance of the network on touching and non-touching entities. It is common knowledge that object detection networks struggle to perform well when entities on the image touch. The hypothesis is proven correct by both Mask RCNN and YOLO as the metric scores are significantly lower for images with touching entities in comparison to the ones that don't. Similar behavior is observed on both the synthetic and real seed image test data sets. It is also worth pointing out that the networks perform phenomenally on images with only non-touching entities achieving recall and precision of ~1.0. It is also observed that the performance of the models on the real-world data sets is OK but not close to that of the performance on synthetic datasets that they are trained on.

The metrics to evaluate the models are Recall and Average Precision (AP) which are the standard evaluation metrics for object detection and localization models such as Mask RCNN and YOLO. Recall is defined as the true-positive rate or the ratio of true positives detected by the model to the total number of objects present. In object detection and localization networks, recall is generally defined over different thresholds of a parameter called Intersection-over-Union (IOU). For the purposes of this work, recall is defined on the IOU of the bounding rectangles predicted by the network and the ground-truth bounding rectangles. The AP of each class is computed using IOU of masks predicted by the network and the ground-truth masks. As pointed out in the work by Yoda et al., the use of bounding rectangles is not an accurate measure since the rectangles are not the minimum area bounding rectangles and don't tightly bind around the instances in the image. The use of IOU over masks helps alleviate the problem since the masks resemble the original instance to a higher degree as they represent the orientation of the instance in the image.

The performance of the models along with pre-trained weights used to train the model on each of the networks is shown in table 1.

**Table 1:** Performance of Models on Synthetic and Real Seed Data Sets

| Seed | Model | | Synthetic Data Set | | | Real Seed Data Set | | |
|---|---|---|---|---|---|---|---|---|
| | Network | Weights | Recall$_{50}$ | AP$_{50}$ | AP@[.5: .95] | Recall$_{50}$ | AP$_{50}$ | AP@[.5: .95] |
| Canola | M-RCNN | ImageNet | 0.88 | 0.90 | 0.76 | 0.78 | 0.77 | 0.62 |
| | M-RCNN | COCO | 0.90 | 0.92 | 0.77 | 0.74 | 0.78 | 0.64 |
| | YOLOv5x | COCO | 0.92 | 0.94 | 0.80 | 0.81 | 0.77 | 0.65 |
| | YOLO5l | COCO | 0.90 | 0.92 | 0.78 | 0.78 | 0.74 | 0.62 |
| | YOLO5m | COCO | 0.89 | 0.88 | 0.74 | 0.71 | 0.73 | 0.62 |
| | YOLO5s | COCO | 0.82 | 0.83 | 0.69 | 0.69 | 0.71 | 0.56 |
| | T-YOLOv4 | T-YOLOv4 | 0.71 | 0.73 | 0.65 | 0.56 | 0.61 | 0.49 |

| | | | | | | | |
|---|---|---|---|---|---|---|---|
| Rough Rice | M-RCNN | ImageNet | 0.90 | 0.91 | 0.79 | 0.85 | 0.74 | 0.61 |
| | M-RCNN | COCO | 0.89 | 0.89 | 0.75 | 0.81 | 0.74 | 0.60 |
| | YOLOv5x | COCO | 0.94 | 0.97 | 0.82 | 0.83 | 0.81 | 0.68 |
| | YOLO5l | COCO | 0.92 | 0.92 | 0.80 | 0.77 | 0.77 | 0.61 |
| | YOLO5m | COCO | 0.90 | 0.90 | 0.76 | 0.72 | 0.73 | 0.58 |
| | YOLO5s | COCO | 0.84 | 0.87 | 0.71 | 0.64 | 0.67 | 0.51 |
| | T-YOLOv4 | T-YOLOv4 | 0.76 | 0.81 | 0.72 | 0.68 | 0.63 | 0.52 |
| Sorghum | M-RCNN | ImageNet | 0.89 | 0.90 | 0.77 | 0.80 | 0.79 | 0.66 |
| | M-RCNN | COCO | 0.87 | 0.89 | 0.75 | 0.78 | 0.80 | 0.67 |
| | YOLOv5x | COCO | 0.93 | 0.93 | 0.80 | 0.84 | 0.80 | 0.72 |
| | YOLOv5l | COCO | 0.91 | 0.90 | 0.78 | 0.81 | 0.77 | 0.64 |
| | YOLOv5m | COCO | 0.89 | 0.88 | 0.75 | 0.74 | 0.72 | 0.60 |
| | YOLOv5s | COCO | 0.89 | 0.84 | 0.74 | 0.68 | 0.66 | 0.55 |
| | T-YOLOv4 | T-YOLOv4 | 0.73 | 0.78 | 0.66 | 0.67 | 0.70 | 0.56 |
| Soy | M-RCNN | ImageNet | 0.87 | 0.86 | 0.74 | 0.86 | 0.73 | 0.68 |
| | M-RCNN | COCO | 0.88 | 0.89 | 0.76 | 0.84 | 0.76 | 0.63 |
| | YOLOv5x | COCO | 0.94 | 0.91 | 0.77 | 0.88 | 0.77 | 0.66 |
| | YOLOv5l | COCO | 0.88 | 0.88 | 0.73 | 0.81 | 0.71 | 0.62 |
| | YOLOv5m | COCO | 0.87 | 0.88 | 0.71 | 0.82 | 0.66 | 0.63 |
| | YOLOv5s | COCO | 0.83 | 0.81 | 0.65 | 0.74 | 0.64 | 0.53 |
| | T-YOLOv4 | T-YOLOv4 | 0.81 | 0.84 | 0.69 | 0.76 | 0.77 | 0.63 |
| Wheat | M-RCNN | ImageNet | 0.89 | 0.88 | 0.76 | 0.84 | 0.75 | 0.61 |
| | M-RCNN | COCO | 0.89 | 0.86 | 0.74 | 0.81 | 0.71 | 0.64 |
| | YOLOv5x | COCO | 0.91 | 0.93 | 0.79 | 0.82 | 0.74 | 0.63 |
| | YOLOv5l | COCO | 0.89 | 0.92 | 0.77 | 0.80 | 0.72 | 0.62 |
| | YOLOv5m | COCO | 0.86 | 0.90 | 0.75 | 0.82 | 0.67 | 0.58 |
| | YOLOv5s | COCO | 0.81 | 0.82 | 0.71 | 0.76 | 0.66 | 0.57 |
| | T-YOLOv4 | T-YOLOv4 | 0.73 | 0.77 | 0.61 | 0.61 | 0.65 | 0.51 |

**Recall$_{50}$** – Recall values at the bounding box IOU threshold of 50%    **AP$_{50}$** – Average precision values at the mask IOU threshold of 50%

**AP@[.5:.95]** – Mean of AP value from IOU of 50% to 90% with a step size of 5%    **T-YOLOv4** – Tiny-YOLOv4

From the results, it is observed that,

1. Although the synthetic test data set is created from seeds entirely different than those used for training, the metrics for recall and average precision are significantly higher for the synthetic data set in comparison to the real seed data set.

2. The models trained on both the COCO and ImageNet for Mask R-CNN data sets produce similar results on both the synthetic and the real seed test data sets showing that both the pre-trained weights are as good as each other.

3. Of the YOLO models, YOLO5x outperforms the others in both recall and average precision for each one of the seeds. It is inferred that the accuracy of the network is directly influenced by the number of layers in the network.

4. The performance of Tiny-YOLOv4 lags behind the larger Mask R-CNN and YOLO models, except for YOLOv5s on occasion. In all fairness, it is not reasonable to compare it against the larger models considering it is smaller in size architecturally. However, it is worth reiterating that they are fit for mobile applications where the models are required to be compact.

## 4 Morphometry Estimation

The focus of the work from this point shifts towards morphometric estimations of the detected seeds. As part of the morphometric measurements, a method which closely follows the standards laid out by the International Seed Morphology Association, an organization that promotes research in the areas of seed morphology and identification, to estimate the length, width and area of each of the seeds along with the count of seeds on the image is proposed. The proposed technique is dependent upon drawing smallest rotated rectangles around observed contours and works well on both Mask R-CNN and YOLO models for all of the seed types in question. The caveat is that the estimations are dependent solely upon the quality of instance detections made by the networks. In case the network doesn't detect an instance or predicts contours incorrectly, which is not uncommon from the results obtained in section 3, the estimates suffer a lack of accuracy. In order to estimate the morphometry i.e. the area, length and width of each of the seeds in question, the following strategy using OpenCV2 is proposed:

1. Capture an image with multiple US pennies. For this experiment, four pennies are used. It is important to have multiple pennies because it is observed from prior research that the position of the objects on the image impacts their perceived area.

2. Input the image to the trained neural network model and detect the contours of each of the coins on the image and determine the average area of all of the detected coins in metric units. The computation of average area is performed by plotting a rectangle around each of the coin contours that are detected and applying the area of the rectangle in pixels to the area of the coin in $mm^2$. It is known that the diameter of the US penny is 19.05 mm resulting in an area of 284.87 $mm^2$ approximating it as a circle.

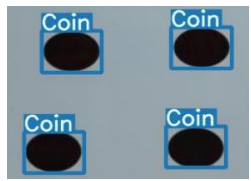

**Figure 5:** Coin Detection by YOLOv5x

3. **(Length and Width Estimation)** Likewise, on images with seeds, detect the contours around each of the seeds in the image and plot the smallest rotated rectangle that encloses each of the contours. The longer axis returned is deemed length and the shorter axis, width.

4. **(Area Estimation)** The area of each of contours is given by a cross-multiplication using the contour area of the detected seed in pixels, contour area of the penny in pixels and area of the penny in $mm^2$ as $seed\ area\ in\ mm^2 = (seed\ area\ in\ pixels * coin\ area\ in\ mm^2)/ (median\ coin\ area\ in\ pixels)$.

Figure 6 shows the detected instances of sorghum seeds on Mask RCNN along with the smallest enclosing rectangles plotted for size estimation using OpenCV2.

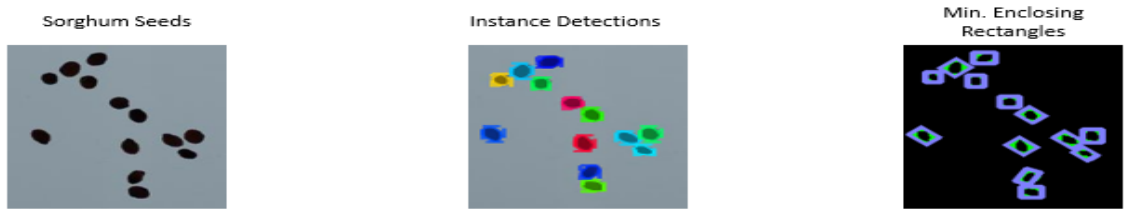

Figure 6: Size Estimation Operations on Sorghum Seeds on Mask R-CNN

## 4.1 Estimated Morphometry in Comparison to Other Applications

The morphometric estimations made by the neural networks are compared against the estimations made mobile applications Leaf-IT and Seed Counter which are in the same realm of morphometry estimation. In addition, the estimations are also compared against the estimations performed by the Mesh algorithm and manual measurements using a vernier caliper. Soy is considered to evaluate for the reason that not all of the seeds in question are able to be estimated by all of the applications in consideration. The areas estimated by each of the them on a sample of seven soy seeds is as shown in Figure 7. Please note that the experiment is repeated on all of the applications with the seed maintained at the same orientation in order to cut out any bias due to orientation and that the seeds do not touch on the images. The type of mesh used for the application of the Mesh algorithm is hexagonal.

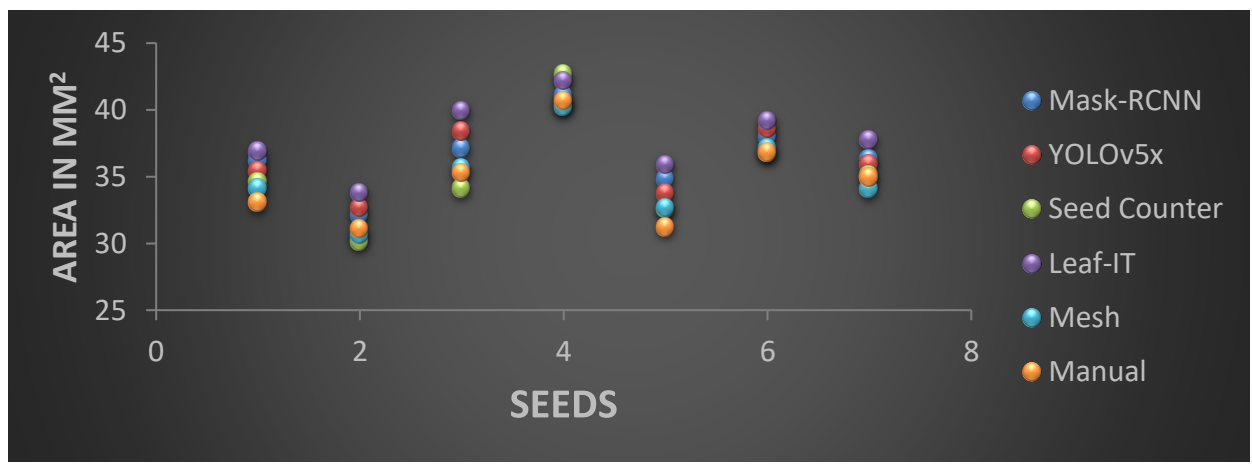

Figure 7: Seed Area Estimates using Morphometry Estimation Applications

The results show a high degree of cadence amongst the estimates made by each of the applications. This proves that the estimates made using neural network detections coincide the applications that currently prevail.

## 5. Conclusion

The work presents an insight into the use of state-of-the-art neural networks of Mask R-CNN and YOLO for seed phenotyping using domain randomization and transfer learning. The findings of the study show that the use of neural networks is a feasible solution for seed detection and morphometry estimation. While the feasibility of the techniques is established, it is important to note that there is room for improvement in terms of accuracy and translation of models from the synthetic to real-world data sets.

## Conflict of Interest

The authors declare no conflict of interest.

## Data and Code Availability

The experiments are performed on Google Colab using the pro subscription to ensure uninterrupted and greater availability of the GPU. A Tesla V100 GPU is used. The experiments for Mask R-CNN and YOLOv5 are run for a total of 40 iterations and Tiny-YOLOv4 is run for a total of 4000 steps. The code and part of the images for the work are made available at

https://github.com/VenkatMargapuri/SeedPhenotyping. Any interest in the work is highly appreciated and the authors are willing to collaborate to expand upon the work.